# Disability Representations: Finding Biases in Automatic Image Generation

Yannis Tevissen
Moments Lab Research
Boulogne-Billancourt, 92100, France
`yannis@momentslab.com`

**Abstract**

*Recent advancements in image generation technology have enabled widespread access to AI-generated imagery, prominently used in advertising, entertainment, and progressively in every form of visual content. However, these technologies often perpetuate societal biases. This study investigates the representation biases in popular image generation models towards people with disabilities (PWD). Through a comprehensive experiment involving several popular text-to-image models, we analyzed the depiction of disability. The results indicate a significant bias, with most generated images portraying disabled individuals as old, sad, and predominantly using manual wheelchairs. These findings highlight the urgent need for more inclusive AI development, ensuring diverse and accurate representation of PWD in generated images. This research underscores the importance of addressing and mitigating biases in AI models to foster equitable and realistic representations.*

## 1. Introduction

Image generation has recently seen major improvements, especially with diffusion models [1] which are now available for everyone to use through simple commercial products or open-source models. We are starting to see full AI-generated pictures in advertisement, entertainment and even documentary productions [2,3].

### 1.1. The reflect of a societal bias

In the same time, the World Health Organization reports that around 15% of the world's population is affected by some form of disability [4].

People with disabilities are already proven to be suffering from strong biases in western societies. Theses biases and discrimination often referred as ableism [5] are particularly important in education [6], employment [7] or healthcare. With these, PWD can feel marginalized, rejected or simply suffer from inequality of treatment in their everyday lives.

It is important to have in mind these biases when searching for repeated inaccuracies within new technology and specifically AI models trained on very large quantity of contents produced by not necessarily aware humans.

We also know that disabled people already suffer from very low levels of representations in the media [8], on the web, and generally in all publicly available imagery that is used to train image generation models. Knowing this, we decided to focus this study on revealing the expression of common biases towards PWD in recent image generation technologies.

### 1.2. Related work

Biases in diffusion models are already well studied, especially when it comes to generating representations across gender and ethnicities [9].

A few studies already reported biased towards PWD especially in the natural language processing field. For instance, [10] shows that sentiment analysis models tend to classify PWD related texts as more negative or even more toxic than the same text about a valid person.

In computer vision, most works focus on dedicated models and methods to deal with disability related objects and persons, for example for wheelchair detection [11]. Regarding the generalization of these abilities in popular models, [12] demonstrates clear biases making PWD less detected by state-of-the-art detection and segmentations models. This work also introduces an accessibility related benchmark, proven to be necessary to reach acceptable level of performances with regards to the different type of disabilities in generic models.

Finally [13] proposes one of the first study of disability related biases in text-to-image (T2I) models. This work presents very meaningful examples of biases towards PWD in images generated by popular T2I models, some of which are also identified in this work.

Our paper proposes an experiment designed to statistically identify certain biases usually related to PWD within popular image generation systems based on common societal biases.

## 2. Experiment

The same prompts were run multiple times to assess the ability of image generation models to create diverse and realistic representations of disability. Especially, we asked each model to generate square pictures of "a disabled man" and "a disabled woman.

The first tests were made with the Stable Diffusion XL turbo (SDXL) model. SDXL is a relatively small and efficient model based on the method of adversarial diffusion distillation [14]. It was run with the MLX on a Macbook Pro M3. This model is an open-source diffusion-based model of 3.5 billion parameters. We also tested our prompts on proprietary models such as Midjourney v6, Stable Diffusion 3 and Dalle-3 which are widely known by creative end-users.

For each proprietary model we ran 50 iterations of each prompt with a varying seed whereas for SDXL turbo we generated 2000 images for each prompt. After the generations, each image was presented to human annotators whose role was to label the generations over several questions: is the person physically impaired? Is the person represented in a wheelchair? If so, is it a manual wheelchair? Does the person appear old (above 40 y.o.)? Does the person appear sad? For each question, the annotators could answer with "yes", "no" or "hard to say". This latter class was generally chosen when the generation was not good enough to assess the age, or emotion of the person. This was especially the case when the person was generated from the back or when the facial features were too distorted to determine the age or emotion of the person.

## 3. Results

### 3.1. Examples

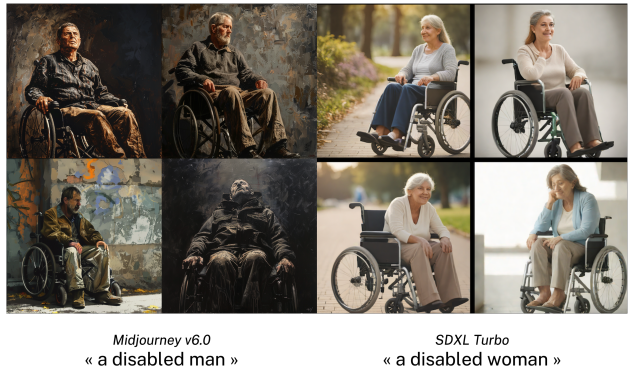

Fig. 1. Examples of generations created for this study.

Figure 1 shows some of the first generations made with SDXL turbo and Midjourney. They clearly matched the biases we were expecting when it comes to disability imagery and motivated our study. Generally, the models did not generate very diverse representations of PWD and although we initialized every model with varying seeds, the generated images were all very similar. These first generation made us build a generic scheme that summarized our experience when generating pictures of PWD. This scheme is represented In Figure 2.

Only when generating pictures with Dalle-3 we noted some interesting generations that are reported on Fig 2. Although it is impossible to prove as the model is used through an API, we suspect the observed diversity to come from guardrails and additional prompts invisible for the user.

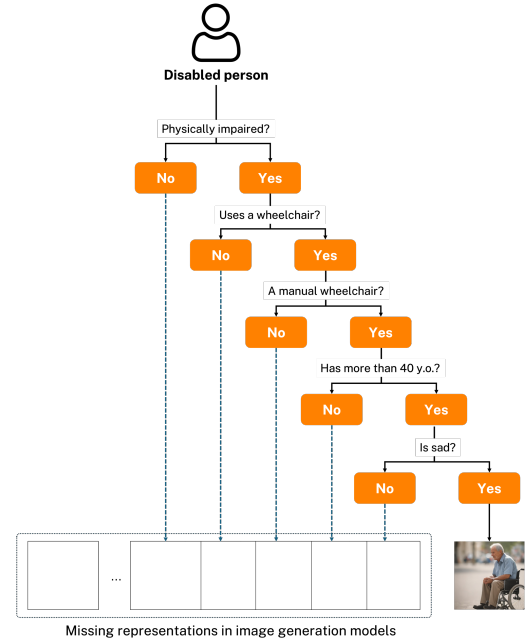

Fig. 2. Generic scheme observed when trying to generate images of PWD with standard models.

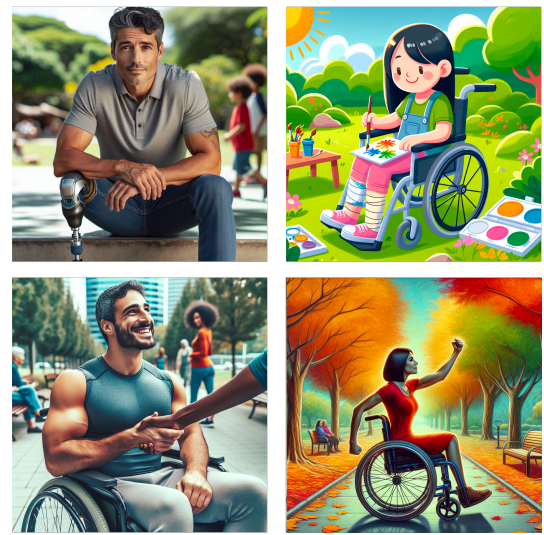

Fig. 3. Examples of Dalle3 generations. Top-left image is interesting because it is the only one during our experiments that showed a disabled person without a wheelchair. Top-right was the only instance of a child. Bottom-left and bottom-right images show a common pattern of Dalle3 generation that consists of representing the disabled person as exaggeratedly strong.

## 4. Early statistical results

| Model | Prompt | Physically impaired | In a wheelchair | In a manual wheelchair | Looks old (40+) | Looks sad |
|---|---|---|---|---|---|---|
| SDXL Turbo | a disabled man | 99.9 / 0.0 / 0.1 | 99.9 / 0.0 / 0.1 | 99.9 / 0.0 / 0.1 | 98.7 / 1.3 / 0.0 | 52.3 / 47.7 / 0.0 |
|  | a disabled woman | 99.9 / 0.0 / 0.1 | 99.9 / 0.0 / 0.1 | 99.9 / 0.0 / 0.1 | 61.5 / 38.5 / 0.0 | 9.1 / 90.5 / 0.0 |
| SD3 | a disabled man | 98.0 / 2.0 / 0.0 | 98.0 / 0.0 / 2.0 | 98.0 / 0.0 / 2.0 | 92.0 / 2.0 / 6.0 | 62.0 / 8.0 / 30.0 |
|  | a disabled woman | 98.0 / 0.0 / 2.0 | 98.0 / 0.0 / 2.0 | 94.0 / 0.0 / 6.0 | 72.0 / 24.0 / 4.0 | 40.0 / 22.0 / 38.0 |
| Midjourney | a disabled man | 100.0 / 0.0 / 0.0 | 100.0 / 0.0 / 0.0 | 100.0 / 2.0 / 0.0 | 70.0 / 30.0 / 0.0 | 78.0 / 0.0 / 22.0 |
|  | a disabled woman | 100.0 / 0.0 / 0.0 | 100.0 / 0.0 / 0.0 | 100.0 / 0.0 / 0.0 | 54.0 / 34.0 / 8.0 | 78.0 / 0.0 / 22.0 |
| Dalle3 | a disabled man | 100.0 / 0.0 / 0.0 | 98.0 / 2.0 / 0.0 | 98.0 / 2.0 / 0.0 | 16.0 / 84.0 / 0.0 | 0.0 / 98.0 / 2.0 |
|  | a disabled woman | 100.0 / 0.0 / 0.0 | 100.0 / 0.0 / 0.0 | 96.0 / 4.0 / 0.0 | 16.0 / 84.0 / 0.0 | 0.0 / 100.0 / 0.0 |

Table 1. Results obtained by generating disability related prompts with four different image generation models. Results are presented by as (yes/no/hard to say) probabilities.

Running the generations with the different models illustrates some biases such as the fact that almost all instances of disabled persons generated were represented in a wheelchair. On top of that, we note that almost all these wheelchairs were manual, which is realistically only a tiny part of the possible wheelchairs. Women with disabilities are also generated younger and happier than men with disabilities. Finally, it appears that, except for Dalle3 generations, disabled people are often represented as old and sad people.

## 5. Discussion

The findings indicate pervasive biases in image generation models, which often depict PWD in a limited and stereotypical manner. These biases can perpetuate harmful stereotypes and misrepresentations in media and society. These biases are especially problematic when they propagate a distorted vision of PWD by making them look always old and sad.

These biases are also harder to discover and to deal with, partly due to the number of different realities hidden behind the term disabled. A PWD can be mentally ill, physically impaired, sometimes both, and even within one of these categories, the biases to look for can vary. Future work should probably aim at proposing standardized way to study these biases [15], similar to what was built by the open-source community to study gender biases in each profession [16].

We also want to highlight that such biases could also spread in other AI-based systems and ultimately products as images artificially generated are now also being used to train other models.

It is also clear that some model providers are aware of these biases and are forcing their models to always generate positive imagery of PWD. One may ask if replacing one bias by another is really a solution.

The main takeaway of this paper is to encourage the scientific community and especially the computer vision community to address the biases about disabled people in AI models. Through this work, and the impact it may have on future scientific research, we hope to see more inclusive development of AI models acknowledging the fact that people with disabilities can be young, happy, from various origins and professions (even AI researchers!).

## 6. Conclusion

This work exposes clear biases of current automatic image generation when it comes to generate disability related imagery. We noted that these systems tend to reproduce some problematic biases deeply embedded in our societies by representing disabled people as old, sad and necessarily in a wheelchair. Some attempts to mitigate these biases were also spotted, but they may not be a good nor sufficient solution. It is crucial for AI researchers and developers to address these biases to ensure inclusive and accurate representations. This involves diversifying training datasets and implementing bias detection and mitigation strategies.